\documentclass[sigconf]{acmart}
\AtBeginDocument{%
  }

\usepackage{soul}
\usepackage{url}
\usepackage{graphicx}
\usepackage{amsmath}
\usepackage{amsthm}
\usepackage{booktabs}
\usepackage[switch]{lineno}
\usepackage{mdframed}
\usepackage{bbding}
\usepackage{pifont}
\usepackage[ruled]{algorithm2e}
\usepackage[export]{adjustbox}
\usepackage{threeparttable}
\usepackage{subcaption}
\usepackage{multirow}
\usepackage{enumitem}
\usepackage{newfloat}
\usepackage{listings}
\usepackage{colortbl}
\usepackage{arydshln}
\usepackage{amsfonts}
\usepackage{wrapfig}
\usepackage{pifont}
\usepackage{float}
\usepackage{bbm}
\usepackage{bm}
\usepackage{mathrsfs}  
\usepackage{makecell}
\usepackage{amsmath}
\usepackage{tabularx}
\usepackage{adjustbox}
\usepackage{graphicx}
\usepackage{booktabs} 
\usepackage{array} 

\usepackage[most]{tcolorbox} 

\definecolor{gray}{rgb}{0.3, 0.3, 0.3}

\newcolumntype{Y}{>{\centering\arraybackslash}X}

\newcolumntype{x}[1]{>{\centering\arraybackslash}p{#1pt}}
\newcolumntype{I}{!{\vrule width 1pt}}

\makeatletter
\newcommand{\thickhline}{%
    \noalign {\ifnum 0=`}\fi \hrule height 1pt
    \futurelet \reserved@a \@xhline
}

\newcommand{\ourmethod}{{\fontfamily{lmtt}\selectfont \textbf{I2P}}\xspace}

\usepackage{hyperref}

\usepackage{amsmath}
\usepackage{mathtools}
\usepackage{amsthm}

\usepackage{balance}

\usepackage[capitalize,noabbrev]{cleveref}
\crefname{figure}{Figure}{Figures}
\Crefname{table}{Table}{Tables}

\setcopyright{acmlicensed}

\copyrightyear{2026}
\acmYear{2026}
\setcopyright{cc}
\setcctype{by}
\acmConference[MM '26]{Proceedings of the 34th ACM International Conference on Multimedia}{November 10--14, 2026}{Rio de Janeiro, Brazil}
\acmBooktitle{Proceedings of the 34th ACM International Conference on Multimedia (MM '26), November 10--14, 2026, Rio de Janeiro, Brazil}
\acmDOI{10.1145/3767308.3836052}
\acmISBN{979-8-4007-2213-4/2026/11}

\begin{document}

\title{Adaptive Forensic Feature Refinement via Intrinsic Importance Perception}

\author{Jiazhen Yang}
\affiliation{%
  \institution{Zhejiang University}
  \city{Hangzhou}
  \country{China}
}
\email{jiazhenyang@zju.edu.cn}

\author{Junjun Zheng}
\affiliation{%
  \institution{Taobao \& Tmall Group of Alibaba}
  \city{Hangzhou}
  \country{China}
}
\email{fangcheng.zjj@alibaba-inc.com}

\author{Kejia Chen}
\affiliation{%
  \institution{Zhejiang University}
  \city{Hangzhou}
  \country{China}
}
\email{chenkejia@zju.edu.cn}

\author{Xiangheng Kong}
\affiliation{%
  \institution{Taobao \& Tmall Group of Alibaba}
  \city{Hangzhou}
  \country{China}
}
\email{yongheng.kxh@alibaba-inc.com}

\author{Jie Lei}
\affiliation{%
  \institution{Zhejiang University of Technology}
  \city{Hangzhou}
  \country{China}
}
\email{jasonlei@zjut.edu.cn}

\author{Zunlei Feng}
\affiliation{%
  \institution{Zhejiang University}
  \city{Hangzhou}
  \country{China}
}
\email{zunleifeng@zju.edu.cn}

\author{Bingde Hu}
\affiliation{%
  \institution{Zhejiang University}
  \city{Hangzhou}
  \country{China}
}
\email{tonyhu@zju.edu.cn}

\author{Yang Gao}
\authornote{Corresponding author.}
\affiliation{%
  \institution{Zhejiang University}
  \city{Hangzhou}
  \country{China}
}
\email{roygao@zju.edu.cn}

\renewcommand{\shortauthors}{Jiazhen Yang et al.}

\begin{abstract}
      With the rapid development of generative models and multimodal content editing technologies, the key challenge in synthetic image detection (SID) lies in cross-distribution generalization to unknown generation sources. Visual foundation models (VFM), which acquire rich visual priors through large scale image-text alignment pretraining, have become a promising approach to improving SID generalization. However, existing VFM-based methods remain coarse-grained in their adaptation strategies. They typically either directly use the final layer representations of VFM or simply fuse multi layer features, lacking explicit modeling of the optimal representational hierarchy for transferable forgery cues. Although fine-tuning VFM can enhance task adaptation, it may also damage the cross-modal pretrained structure that supports open-set generalization. To address this task specific tension, we reformulate VFM adaptation for SID as a joint optimization problem: it is necessary both to identify the critical representational layer that is more suitable for carrying forgery discriminative information and to constrain the disturbance caused by task knowledge injection to the pretrained structure. Based on this, we propose \ourmethod{}, an SID framework centered on intrinsic importance perception. \ourmethod{} first adaptively identifies the critical layer representations that are most discriminative for SID, and then constrains task-driven parameter updates within a low sensitivity parameter subspace, thereby improving task specificity while preserving the transferable structure of pretrained representations as much as possible. We conduct systematic experiments and ablation analyses on multiple cross-domain benchmarks, and the results show that \ourmethod{} achieves stable and competitive generalization performance. 
\end{abstract}



\maketitle

\begin{figure*}
    \centering
    \includegraphics[width=\textwidth]{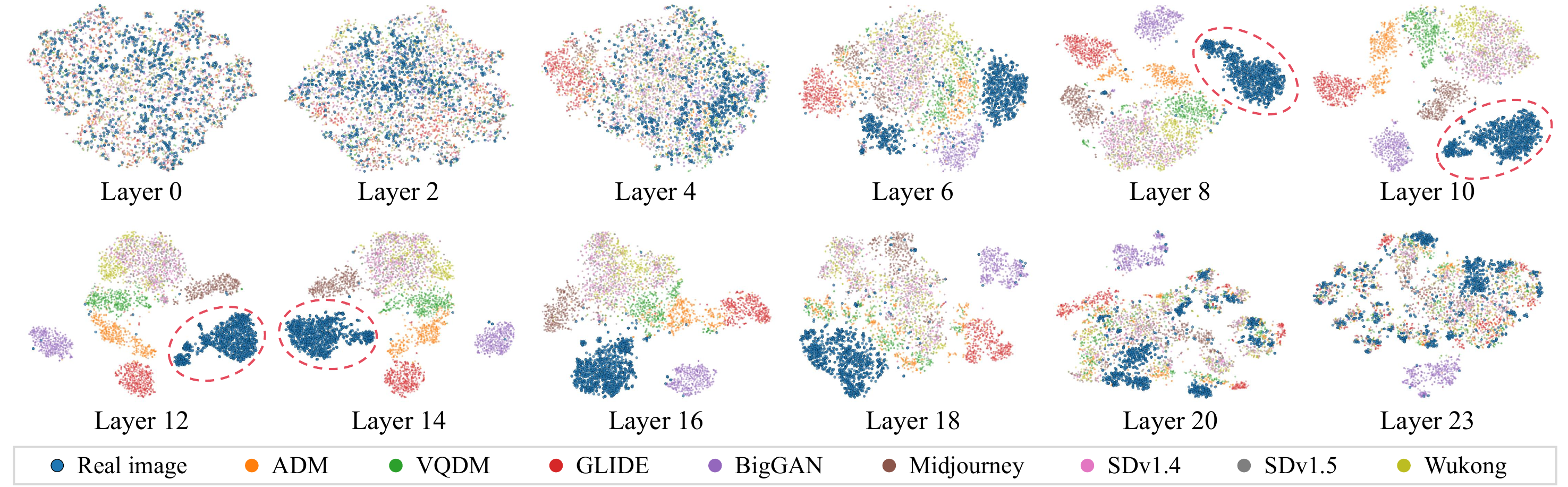}
    \caption{\textbf{T-SNE visualizations of layer-wise representations extracted on GenImage~\cite{zhu2023genimage} using a frozen CLIP ViT-L/14~\cite{radford2021learning} visual encoder.} Intermediate layers naturally exhibit a clearer separable structure between real and synthetic embeddings, whereas in early and deep stages the feature space overlap across categories becomes more pronounced.}
    \label{fig:banner}
\end{figure*}

\section{Introduction}
Rapid advances in AI-generated imagery and multimodal editing have lowered the barrier to producing high-fidelity synthetic content~\cite{wang2025dfbench, zhang2025unified, ho2020denoising, midjourney2022}, whose large-scale dissemination raises growing social risks~\cite{cao2025survey}. Accordingly, reliable synthetic image detection (SID) has become increasingly important~\cite{zhang2025asap, shen2025beyond, shencore, shenomnivl}. The central challenge lies in \textbf{cross-domain generalization}: the samples to be detected in real-world often come from generation sources not covered during training, which requires detection models to generalize stably to unseen generative models under limited supervised data.

Most specialized detection methods rely on systematic artifacts left during the generation process, using deep networks to capture cues such as handcrafted noise patterns~\cite{yan2024sanity, liang2025ferretnet}, frequency statistics~\cite{tan2024frequency, zhang2025frequency}, and local correlations~\cite{tan2024rethinking, li2025improving}. They typically require large and diverse training data for high accuracy. When faced with unknown post-processing or new forgery types, their dependence on pixel-level artifacts often leads to significant performance degradation. More importantly, this paradigm requires continual retraining to keep pace with evolving generative models, leaving detection systems passive and lagging in practical deployment.

Recent studies have shown that pre-trained visual foundation models (VFM) ~\cite{awais2025foundation}, which learn general-purpose visual representations from large-scale datasets, have greater potential to improve the generalization capabilities of SID and reduce reliance on training distribution coverage compared to specialized detectors that heavily rely on specific artifact patterns.
However, existing VFM-based SID transfer strategies remain coarse-grained. First, many methods treat VFM as fixed feature extractors, directly using the final layer representations~\cite{zhang2025towards, ojha2023towards} or simply fusing multi layer features~\cite{chen2025forgelens, koutlis2024leveraging}, while overlooking the functional differentiation of hierarchical representations and the potential suppression of fine-grained forgery cues by deep semantic compression. Second, although parameter fine-tuning~\cite{huang2025keeping} can enhance the task adaptability of VFM~\cite{yan2024effort, liu2024forgery}, it may still induce representational drift under limited downstream supervision, thereby damaging the pretrained structure that supports generalization. 
Therefore, fully unlocking the generalization potential of VFM for SID hinges on two interrelated questions: \textbf{I)} \textit{How to identify the critical representational hierarchy most effective to forgery discrimination.} \textbf{II)} \textit{How to inject task knowledge while minimizing disturbance to the pretrained structure.}

To this end, we propose a task-oriented adaptation framework, adaptive forensic feature refinement via \textbf{I}ntrinsic \textbf{I}mportance \textbf{P}erception, termed \ourmethod{}, to better explore the transfer potential of VFM for SID task. Our key observation is that hierarchical representations in VFM contribute unevenly to SID. As shown in ~\cref{fig:banner}, with VFM parameters frozen, intermediate layers naturally exhibit discriminative structured representations, while shallow layers are more susceptible to low-level confounding information and deeper layers weaken fine-grained forgery cues due to stronger semantic aggregation. This suggests that fully exploiting the generalization advantage of VFM in SID requires identifying the critical representational hierarchy that is more suitable for extracting forgery discriminative information, while avoiding damage to the discriminative structures formed in the intermediate layers during adaptation.

Building on this insight, to respond to question \textbf{I)}, we propose Critical Layer Identification (CLI) to adaptively locate the critical layer representation that is most discriminative for SID, mitigating information dilution caused by indiscriminate use of layers. Driven by question \textbf{II)}, we further propose Controlled Knowledge Injection (CKI), which constrains parameter updates driven by downstream supervisory signals within a low-sensitivity parameter subspace, so as to minimize the interference of representational drift with the pretrained transferable structure while improving task adaptability. Through the above design, \ourmethod{} unifies the selection of critical representational hierarchy and controlled task adaptation into a single framework, thereby enhancing the discriminative capability of SID while more robustly preserving the cross-distribution transfer capability of VFM. The main contributions are as follows:

\begin{itemize}[leftmargin=*]

\item[\ding{182}] We systematically analyze the differential roles of hierarchical representations in VFM for SID, and find that the representations more valuable for forgery discrimination do not come from the deepest semantic layers. Meanwhile, we point out that preserving the pretrained transferable structure during adaptation is an important factor affecting cross-distribution generalization.

\item[\ding{183}] We propose \ourmethod{}, which unifies the localization of critical representational hierarchies and controlled knowledge injection into the same adaptation framework. CLI adaptively locates the critical layer representation that is most discriminative for SID, while CKI constrains updates within the low-sensitivity parameter subspace, achieving more robust downstream adaptation.

\item[\ding{184}] We conduct comprehensive experiments and ablation studies on multiple cross-domain benchmarks. The results verify that \ourmethod{} achieves stable and competitive generalization performance on cross-distribution synthetic image detection task.

\end{itemize}

\section{Related Work}
\noindent\textbf{Specialized SID methods.}
Existing SID methods commonly construct discriminative representations around forgery cues in the pixel, frequency, or reconstruction domains. FreqNet~\cite{tan2024frequency}, SPAI~\cite{karageorgiou2025any}, and GLDF~\cite{zhang2025frequency} exploit anomalous frequency responses and spectral structures, while NPR~\cite{tan2024rethinking} and SAFE~\cite{li2025improving} model unnatural local correlations. PatchCraft~\cite{zhong2023patchcraft}, ESSP~\cite{chen2024single}, and FerretNet~\cite{liang2025ferretnet} capture imaging noise and fine-grained textures that are difficult for generative models to reproduce. DIRE~\cite{wang2023dire}, LaRE~\cite{luo2024lare}, and STD-FD~\cite{loustd} characterize generation differences through reconstruction inconsistencies. Although effective under known-generator settings, these methods rely on specific artifacts and often degrade under unseen generators, post-processing, or distribution shifts.

\noindent\textbf{VFM-based SID methods.}
Recent studies introduce visual foundation models into SID to leverage transferable priors learned from large-scale pretraining and reduce dependence on training-distribution coverage. Unlike specialized detectors built around a particular type of forgery artifact, VFM provide more general visual structures and semantic priors learned from diverse data. UniFD~\cite{ojha2023towards} and VIB~\cite{zhang2025towards} use frozen VFM features for forgery classification, whereas FatFormer~\cite{liu2024forgery} and ForgeLens~\cite{chen2025forgelens} employ adaptation modules to emphasize forgery-related cues. Effort~\cite{yan2024effort} improves adaptation through subspace decomposition and selective parameter updating, while RINE~\cite{koutlis2024leveraging} and ForgeLens~\cite{chen2025forgelens} exploit features from multiple hierarchical levels to enhance transferability. Despite demonstrating the potential of pretrained representations, existing methods still lack unified modeling of where transferable forgery cues should be extracted and how task knowledge should be injected without disrupting pretrained structures. In contrast, we emphasize that SID-relevant information is unevenly distributed across layers, and that relying solely on final-layer features or indiscriminately fusing multiple layers may dilute discriminative cues and introduce irrelevant information, thereby limiting cross-distribution generalization.

\begin{figure}
    \centering
    \includegraphics[width=0.48\textwidth]{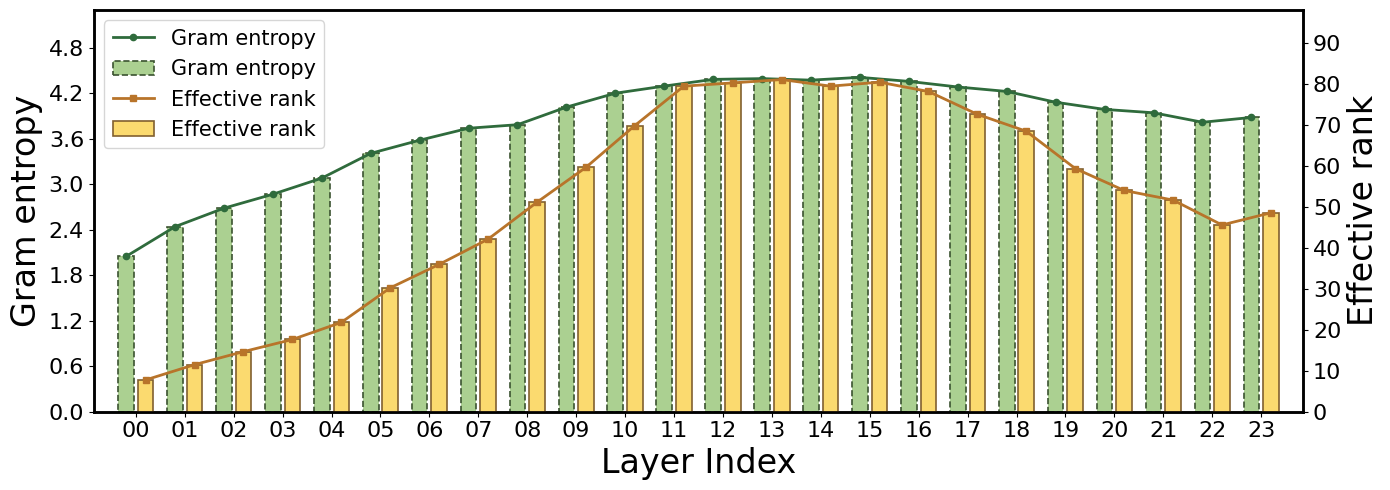}
    \caption{\textbf{Layer-wise spectral structure analysis of a frozen CLIP ViT-L/14.} For each layer’s activations, compute the Gram matrix entropy (\textcolor[HTML]{2f6b3c}{left axis}) and the effective rank (\textcolor[HTML]{b8742a}{right axis}) to quantify spectral energy dispersion and the effective dimensionality.}
    \label{fig:gram_rank}
\end{figure}

\section{Motivation and Analysis}

\subsection{Why Intermediate Layer Matters for SID}

Pretrained VFMs develop hierarchical representations along network depth, progressing from local details to high-level semantic concepts. Although most methods use final-layer features as image representations, recent studies~\cite{bolya2025perception, de2025mysteries} show that the optimal features for downstream tasks do not necessarily arise from the deepest layers. For SID, discriminative evidence often lies in fine-grained local artifacts that are weakly related to high-level semantics. We therefore hypothesize that useful forgery cues are concentrated in the intermediate representations of VFMs. To investigate this, we analyze the layer-wise evolution of representations using the CLIP ViT-L/14 visual encoder~\cite{radford2021learning}.

We further interpret this trend from the perspective of the Information Bottleneck (IB)~\cite{tishby2000information}. Let $x$ denote the input image, $f_\ell \in \mathbb{R}^d$ the representation at layer $\ell$, and $y$ the supervision target. IB seeks a representation that preserves target-relevant information while compressing irrelevant details:
\begin{equation}
\mathcal{L}_{\text{IB}} = I(x; f) - \beta I(f; y),\quad \beta > 0,
\end{equation}
where $I(\cdot;\cdot)$ denotes mutual information and $\beta$ controls the trade-off. As depth increases, the pretraining objective progressively strengthens semantic abstraction while suppressing raw details. For SID, excessive semantic aggregation may weaken local statistical traces important for forgery discrimination. We conjecture that an intermediate layer $\ell^*$ provides the best trade-off between noise suppression and fine-grained information preservation.

\begin{figure}
    \centering
    \includegraphics[width=0.48\textwidth]{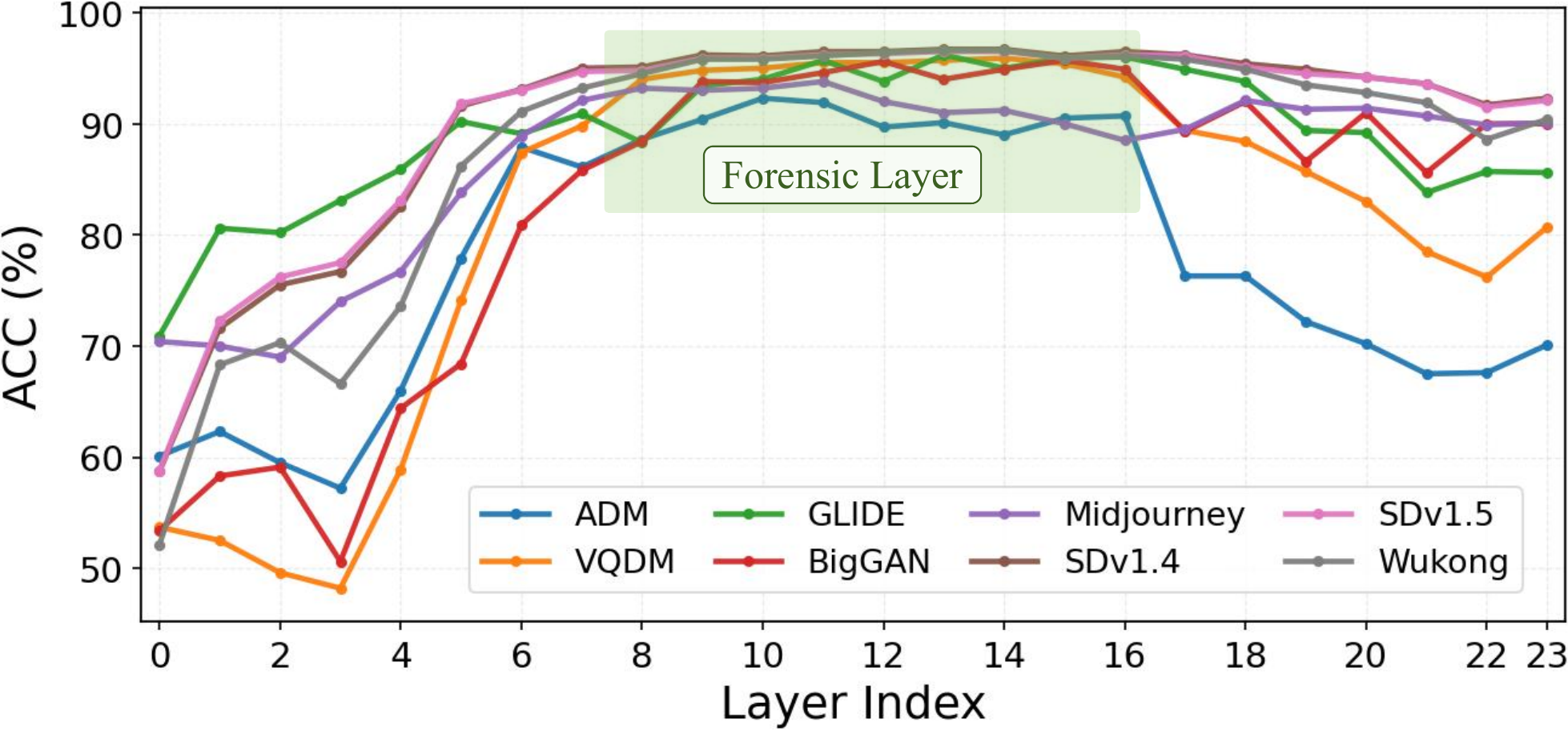}
    \caption{\textbf{Layer-wise linear separability of a frozen CLIP ViT-L/14.} A linear classifier is trained on CLS token features from each layer and evaluated on the GenImage test set, the classification accuracy peaks at intermediate layers.}

    \label{fig:linearprob}
\end{figure}

\begin{figure}
    \centering
    \includegraphics[width=0.48\textwidth]{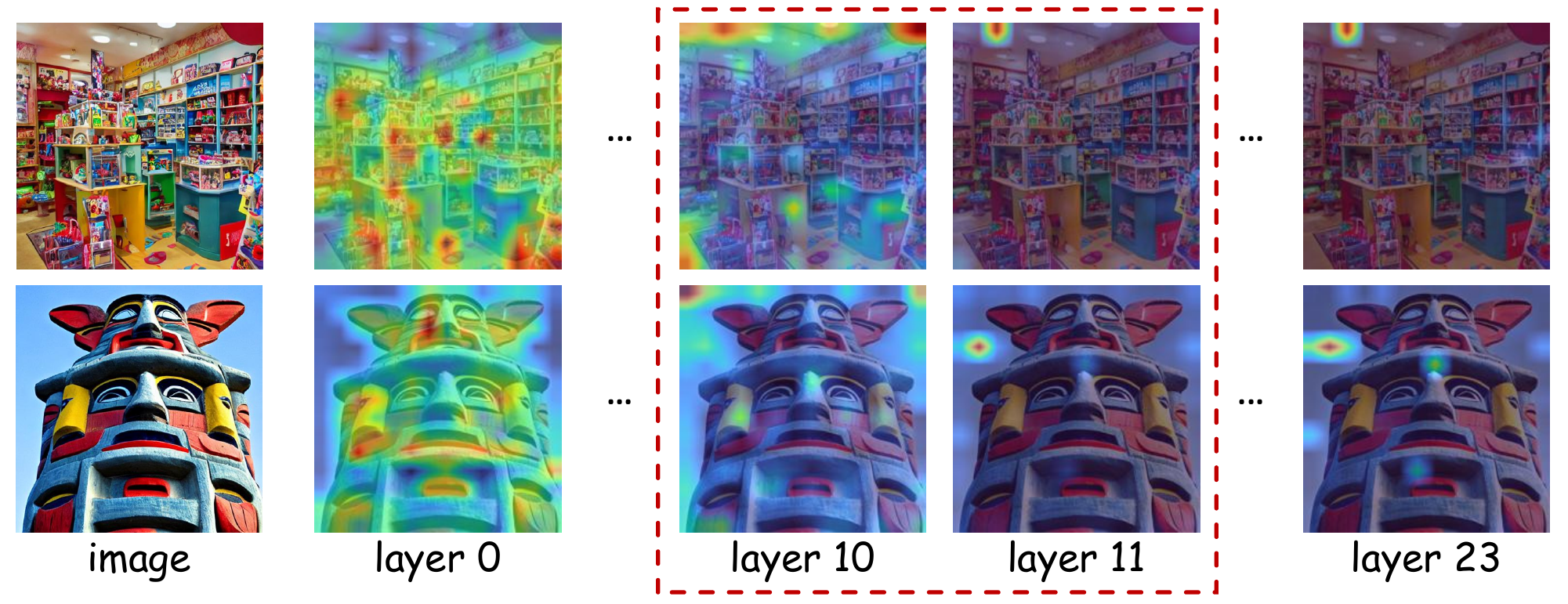}

    \caption{\textbf{Layer-wise evolution of CLS token attention maps in a frozen CLIP ViT-L/14.} Attention is diffuse in early layers, broadly covering local details and textures, while a clear transition emerges around layers 10–11, where the attention rapidly concentrates and remains relatively stable in subsequent layers.}

    \label{fig:attentionmap}
\end{figure}

To provide empirical support for the above conjecture, we analyze layer-wise representations of the frozen visual encoder from the perspective of spectral structure~\cite{skean2025layer, roy2007effective}. Specifically, for the representation $F_\ell \in \mathbb{R}^{n \times d}$ at the $\ell$-th layer, where $n$ is the number of samples, we compute the Gram matrix entropy to characterize the dispersion of spectral energy and use the effective rank to measure the effective dimensionality of the representation. As shown in ~\cref{fig:gram_rank}, the two metrics exhibit consistent layer-wise trends: both reach their peaks around the intermediate layers, indicating that representations at this stage can cover a richer representation space. As depth increases, the spectral energy gradually becomes more concentrated and the effective dimensionality decreases, suggesting that the representations begin to exhibit a more pronounced tendency toward dimensional contraction and semantic compression.

Furthermore, with the encoder frozen, we train a linear classifier $W_\ell \in \mathbb{R}^{d \times 2}$ on the feature $f_\ell$ from each layer separately to evaluate the linear separability of representations from different layers for the SID task. The results in \cref{fig:linearprob} show that the classification accuracy peaks in intermediate layers and then gradually declines in deeper layers. Meanwhile, the CLS token attention visualizations in \cref{fig:attentionmap} indicate that the attention is relatively diffuse in shallow layers, whereas it exhibits a pronounced convergence around intermediate layers and remains stable thereafter. Notably, this transition interval aligns closely with the peak interval in ~\cref{fig:gram_rank}. This suggests that after effective information aggregation is completed in intermediate layers, deeper layers further compress the representation along the direction of semantic alignment.

Based on the above analysis, we draw our key conclusion: within the hierarchical representations of a VFM, there exists a critical intermediate layer $\ell^*$ whose representation forms relatively stable structured features while still preserving fine-grained cues that are valuable for forgery discrimination. Compared with shallow representations that are more susceptible to low-level confounding information and deep representations that are more inclined toward semantic compression, this layer is more suitable as the representational readout location for SID and provides the basis for the subsequent critical layer identification mechanism.

\begin{figure}
    \centering
    \includegraphics[width=0.48\textwidth]{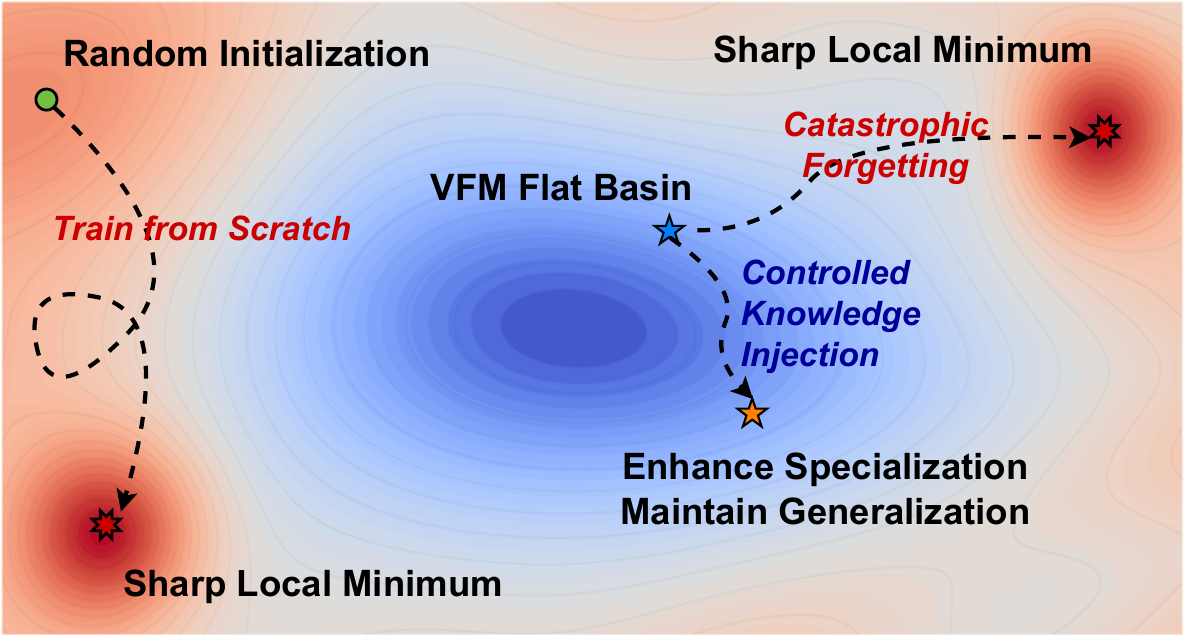}

    \caption{Random initialization (\protect\includegraphics[scale=0.18,valign=c]{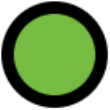}) necessitates extensive exploration and may converge to sharp, perturbation-sensitive local minima (\protect\includegraphics[scale=0.15,valign=c]{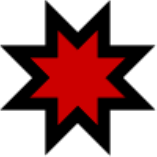}). Pre-trained VFM reside within a structured flat basin (\protect\includegraphics[scale=0.16,valign=c]{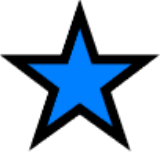}). Aggressive fine-tuning risks escaping this basin and inducing catastrophic forgetting, whereas controlled knowledge injection confines updates to low-sensitivity directions to preserve generalization (\protect\includegraphics[scale=0.16,valign=c]{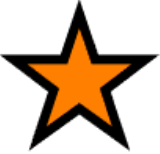}).}

    \label{fig:loss}
\end{figure}

\begin{figure*}
    \centering
    \includegraphics[width=\textwidth]{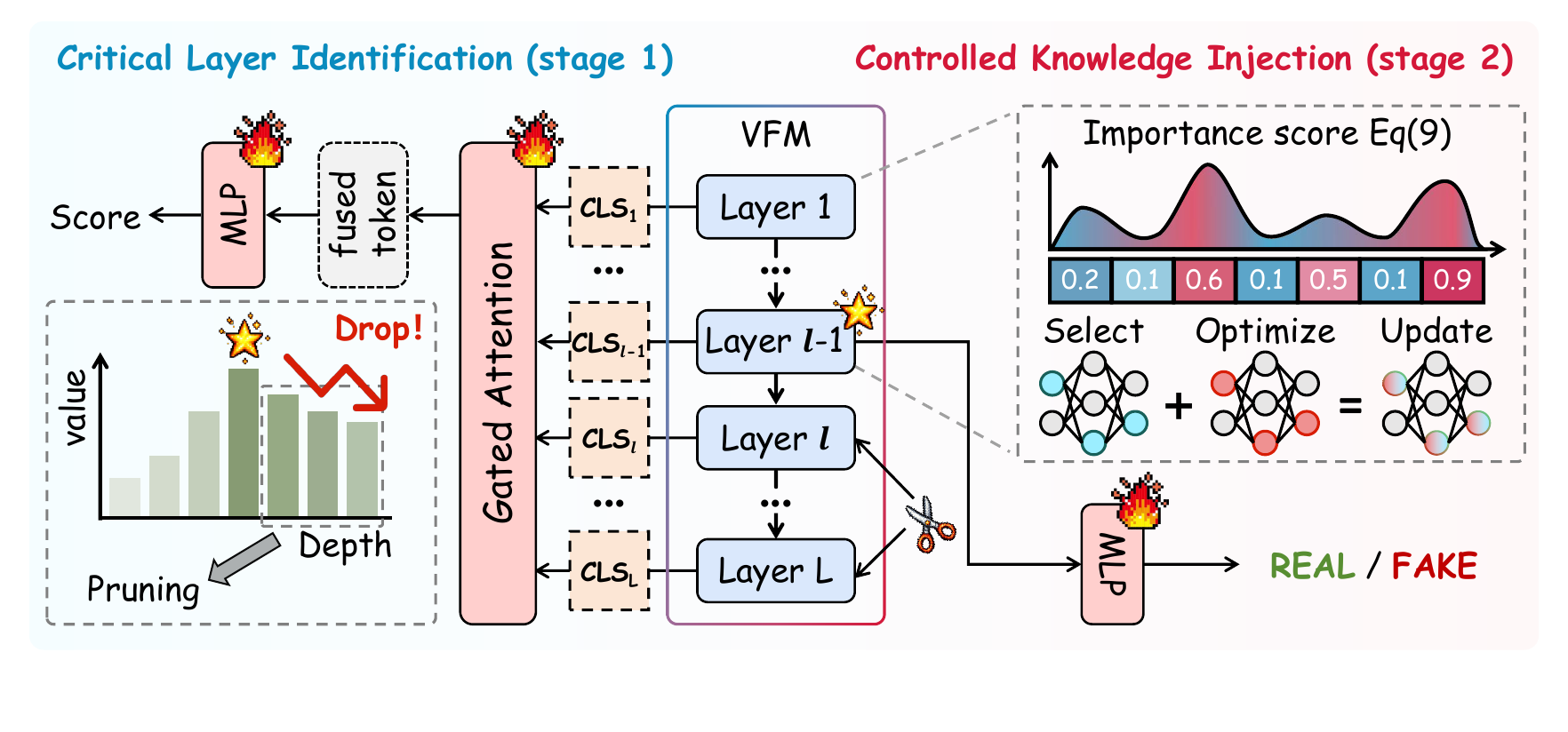}

    \caption{\textbf{\ourmethod{} overall framework.}
    \textbf{Stage 1: \textcolor[HTML]{098BBE}{Critical Layer Identification}} extracts layer-wise CLS token representations from a frozen VFM and learns their relative contributions via gated attention, identifying the most discriminative intermediate layer (\protect\includegraphics[scale=0.7,valign=c]{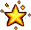}) for SID and pruning (\protect\includegraphics[scale=0.7,valign=c]{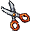}) deeper layers beyond it. \textbf{Stage 2: \textcolor[HTML]{D41738}{Controlled Knowledge Injection}} selects a low-importance parameter subset using importance scores and updates only this subset together with the classification head, enabling controlled injection of forgery-relevant knowledge while maximally preserving the pretrained representation structure.}
    \label{fig:framework}
\end{figure*}

\subsection{Why Pre-trained Knowledge Benefits SID}
\label{sec:CKI}

In the above empirical analysis, we find that even when the CLIP visual encoder is completely frozen, the model can still achieve competitive performance on the cross-generator SID task by relying solely on the critical intermediate-layer features. This indicates that the representations of pretrained VFMs already contain effective structures that can support SID generalization. Therefore, rather than strongly intervening in the representational space, a more reasonable adaptation strategy is to selectively inject downstream task knowledge on the basis of preserving such pretrained structures.

As shown in~\cref{fig:loss}, for a randomly initialized model, its parameters lack the stable visual priors formed through large-scale pretraining. Therefore, model training usually has to rely more heavily on downstream empirical risk minimization, searching for effective solutions from scratch in a high dimensional parameter space. In this case, the optimization process is more easily constrained by the limited training distribution, thereby converging to discriminative boundaries that are more sensitive to data perturbations and distribution shifts. In contrast, a VFM has completed general representation learning on large scale and diverse data $\mathcal{D}_{\text{pre}}$, and its parameters can be regarded as lying near a structured region shaped by pretraining, denoted as $\theta_{\text{pre}}$. This region contains stable visual inductive biases and transferable representational structures, significantly reducing the effective search space of downstream tasks, such that subsequent adaptation often only requires limited adjustment in the vicinity of $\theta_{\text{pre}}$ to obtain an effective solution.

However, if the pretrained VFM is fine-tuned too aggressively under limited downstream supervisory data, the optimization process will follow the empirical risk of the current training distribution more strongly, thereby driving the parameters to deviate significantly from the pretrained structural region corresponding to $\theta_{\text{pre}}$. Such deviation will damage the general representational structure formed during pretraining, making the model more inclined to rely on task-specific patterns tied to the current training distribution, which in turn leads to more pronounced representation drift and affects its generalization ability toward unseen distributions.

Based on the above analysis, we argue that downstream adaptation should not be performed in an unconstrained manner over the entire parameter space. Since different parameter directions do not affect the stability of pretrained representations equally, a more reliable strategy is to prioritize updates within a low sensitivity parameter subspace, so as to improve task adaptation capability while minimizing disturbance to the pretrained transferable structure as much as possible. Specifically, to characterize the local update risk along different parameter directions, we consider a second-order approximation of the downstream loss with respect to the parameter perturbation $\Delta\theta$ in the vicinity of $\theta_{\text{pre}}$:
\begin{equation}
\mathcal{L}(\theta_{\text{pre}}+\Delta\theta)
\approx 
\mathcal{L}(\theta_{\text{pre}})
+\tfrac{1}{2}\,\Delta\theta^\top H\,\Delta\theta,
\end{equation}
where $H=\nabla^2_{\theta}\mathcal{L}(\theta_{\text{pre}})$ denotes the Hessian matrix of the loss function in the vicinity of $\theta_{\text{pre}}$, which describe the curvature characteristics of the local parameter space. This approximation indicates that the influence of parameter perturbations on the loss is governed by the curvature weighted contributions of different directions. Along high curvature directions, even small updates may lead to significant changes in the loss and disrupt pretrained structure stability. In contrast, along low-curvature directions, the optimization process usually takes place in a smoother parameter neighborhood. Therefore, if downstream updates are constrained within a low-curvature, low-sensitivity parameter subspace, it becomes more favorable to enhance task adaptation capability while alleviating damage to the pretrained representational structure and mitigating the risk of generalization degradation caused by over adaptation.

\section{Method}

\subsection{Overview}

The proposed \ourmethod{} framework is illustrated in ~\cref{fig:framework}. It is designed to address two key issues in VFM adaptation for SID: how to extract representations that are more suitable for forgery discrimination, and how to perform downstream adaptation while minimizing disturbance to the pretrained structure as much as possible. To this end, \ourmethod{} consists of two consecutive stages. First, Critical Layer Identification (CLI) adaptively identifies the critical layer that is most discriminative for SID from the hierarchical representations of a frozen VFM. Subsequently, Controlled Knowledge Injection (CKI) constrains the range of downstream updates based on parameter sensitivity estimation, such that task knowledge is primarily injected into a relatively low-sensitivity parameter subspace. This design enhances the discriminative capability of the SID task while preserving the cross-distribution generalization ability of the VFM.

\subsection{Critical Layer Identification}

Given an input image $x$, we first use a frozen visual encoder $\mathcal{F}$ to extract a set of hierarchical representations $f=\{f_\ell\}_{\ell=1}^{L}$ along the model depth $L$. Since different VFM exhibit distinct layer-wise response patterns, and evaluating each layer with handcrafted heuristics is both costly and poorly transferable. Therefore, we propose \textbf{Critical Layer Identification (CLI)} to adaptively identify the critical layer that is most discriminative for SID from the hierarchical representations of a frozen VFM, thereby determining a more appropriate representational readout location.

To characterize the relative contributions of representations from different layers to the SID task, we introduce a lightweight layer scoring function that assigns each candidate representation $f_\ell$ a learnable score:
\begin{equation}
\alpha(f_\ell) = W_2(tanh(W_1f_\ell)+b_1)+b_2,
\end{equation}
where $W_1\in \mathbb{R}^{h \times d}$ and $W_2\in \mathbb{R}^{1 \times h}$ are learnable parameters, $b_1$ and $b_2$ are bias terms, and $tanh(\cdot)$ is the hyperbolic tangent activation function. We then apply softmax normalization to $\{\alpha(f_\ell)\}_{\ell=1}^{L}$ to obtain a layer-wise weight distribution $\{\pi_\ell\}_{\ell=1}^{L}$:
\begin{equation}
\pi_\ell=\frac{\exp(\alpha(f_\ell))}{\sum_{\ell'=1}^{L}\exp(\alpha(f_{\ell'}))}.
\end{equation}

\begin{table*}[t]\small
\caption{Accuracy (ACC, \%) and average precision (AP, \%) comparison of \ourmethod{} and other methods on the Setting 1. All methods are trained on SDv1.4 and evaluated on different test subsets. \textbf{Bold} indicates the best result, and \underline{underline} denotes the second-best.}
\centering
\scriptsize{
\resizebox{\linewidth}{!}{
\setlength\tabcolsep{3.pt}
\renewcommand\arraystretch{1.1}
\begin{tabular}{l|l|cc|cc|cc|cc|cc|cc|cc|cc|cc}
\hline\thickhline
\rowcolor{gray!20}
 &  &
\multicolumn{2}{c|}{Midjourney} &
\multicolumn{2}{c|}{SDv1.4} &
\multicolumn{2}{c|}{SDv1.5} &
\multicolumn{2}{c|}{ADM} &
\multicolumn{2}{c|}{GLIDE} &
\multicolumn{2}{c|}{Wukong} &
\multicolumn{2}{c|}{VQDM} &
\multicolumn{2}{c|}{BigGAN} &
\multicolumn{2}{c}{Average}
\\
\cline{3-20}
\rowcolor{gray!20}
\multirow{-2}{*}{Methods} & \multirow{-2}{*}{Venue}
& ACC. & AP. & ACC. & AP. & ACC. & AP. & ACC. & AP. & ACC. & AP. & ACC. & AP. & ACC. & AP. & ACC. & AP. & ACC. & AP.
\\
\hline

LGard & CVPR 2023
& 73.74 & 77.63 & 76.28 & 79.15 & 77.12 & 80.27 & 51.96 & 51.38 & 49.92 & 50.57 & 73.18 & 75.61 & 52.83 & 51.94 & 40.69 & 39.36 & 61.97 & 63.24
\\
\rowcolor{gray!10} DIRE & CVPR 2023
& 77.43 & 86.49 & 94.42 & 99.09 & 94.13 & 99.09 & 72.08 & 82.18 & 71.23 & 81.65 & 93.99 & 98.85 & 55.83 & 62.69 & 50.66 & 54.58 & 76.22 & 83.08
\\
NPR & CVPR 2024
& 89.84 & 97.61 & 90.74 & 98.77 & 90.74 & 98.82 & 84.63 & 93.16 & 90.31 & 98.34 & 90.67 & 98.06 & 87.04 & 93.05 & 81.84 & 88.64 & 88.23 & 95.81
\\
\rowcolor{gray!10} FreqNet & AAAI 2024
& 69.72 & 78.96 & 64.27 & 74.35 & 64.84 & 75.62 & 83.31 & 91.47 & 81.68 & 88.84 & 57.79 & 66.72 & 81.76 & 89.63 & 90.58 & 94.81 & 74.24 & 82.55
\\
FerretNet & NIPS 2025
& 78.00 & 97.60 & 100.0 & 100.0 & 99.80 & 100.0 & 76.48 & 96.60 & 98.56 & 99.80 & 99.58 & 99.99 & 81.62 & 98.32 & 91.32 & 97.79 & 90.67 & 98.76
\\
\rowcolor{gray!10} UniFD & CVPR 2023
& 72.68 & 92.47 & 85.47 & 97.08 & 85.11 & 97.09 & 55.34 & 67.91 & 77.43 & 93.66 & 76.48 & 93.63 & 57.24 & 78.48 & 81.21 & 96.13 & 73.87 & 89.55
\\
FatFormer & CVPR 2024
& 92.70 & 97.60 & 100.0 & 100.0 & 99.90 & 100.0 & 75.90 & 94.70 & 88.00 & 95.80 & 99.90 & 100.0 & 98.80 & 99.40 & 55.80 & 64.90 & 88.88 & 94.05
\\
\rowcolor{gray!10} RINE & ECCV 2024
& 96.78 & 99.55 & 99.54 & 99.92 & 99.41 & 99.90 & 90.03 & 98.89 & 96.15 & 99.58 & 99.60 & 99.85 & 98.90 & 99.79 & 82.80 & 95.46 & 95.40 & 99.12
\\
ForgeLens & ICCV 2025
& 97.40 & 99.60 & 99.70 & 100.0 & 99.50 & 100.0 & 94.00 & 98.70 & 99.50 & 99.40 & 99.00 & 99.90 & 97.80 & 99.20 & 93.80 & 99.10 & \underline{97.59} & \underline{99.49}
\\
\rowcolor{gray!10} Effort & ICML 2025
& 82.40 & 99.00 & 99.80 & 99.80 & 99.80 & 99.80 & 78.70 & 93.80 & 93.30 & 97.00 & 97.40 & 99.30 & 91.70 & 96.10 & 77.60 & 98.40 & 91.10 & 97.90
\\

\hline
\rowcolor[HTML]{D7F6FF}
\ourmethod{} & -
& 96.86 & 99.30 & 99.72 & 99.99 & 99.63 & 99.98 & 95.02 & 99.38 & 98.80 & 99.78 & 99.46 & 99.93 & 98.63 & 99.78 & 97.64 & 99.59 & \textbf{98.22} & \textbf{99.72}
\\
\hline\thickhline
\end{tabular}}}
\label{tab:table1}
\end{table*}

During the identification stage, we construct an aggregated representation based on this weight distribution:
\begin{equation}
\hat{f}=\sum_{\ell=1}^{L}\pi_\ell f_\ell,
\end{equation}
and feed it into the classification head $g(\cdot)$ to obtain the prediction logits $z$:
\begin{equation}
z = g(\hat{f}).
\end{equation}

The purpose of this process is not to rely on soft fusion for detection in the long term, but to estimate the relative discriminative value of each layer for SID through a learnable layer contribution distribution. After estimating layer-wise contributions, we select the layer with the largest weight as the critical layer $\ell^*$:
\begin{equation}
\ell^*=\arg\max_{\ell}\pi_\ell .
\end{equation}

In the subsequent adaptation and inference stages, we use the representation corresponding to this layer, $f^*=f_{\ell^*}$, as the image representation. In this way, the representational readout for SID is explicitly placed on the single layer with the greatest discriminative value, thereby avoiding the information dilution that may be caused by indiscriminate use of multiple layers. For the deeper modules after $\ell^*$ that no longer participate in representation modeling, structured pruning can be directly applied, thereby avoiding redundant computational overhead. Through the above process, CLI can adaptively identify the critical discriminative layer $\ell^*$ for SID from the hierarchical representations of a frozen VFM, providing a more targeted representational basis for subsequent adaptation.

\subsection{Controlled Knowledge Injection}

After identifying the critical layer, we further consider how to improve the task adaptation capability of the VFM for SID while keeping the pretrained representational structure relatively stable. To this end, we propose \textbf{Controlled Knowledge Injection (CKI)}. The core idea is to use the local sensitivity of parameter directions to constrain the range of downstream updates, such that task-related supervision is primarily injected into a relatively low-sensitivity parameter subspace.

To evaluate the local sensitivity of different parameter directions, we adopt a practical approximation based on second-order information to estimate parameter importance. Considering that explicitly computing the full Hessian is infeasible for large-scale models, for a linear mapping layer $W \in \mathbb{R}^{m \times d}$, we use the second moment of activations to approximate the local curvature~\cite{frantar2023sparsegpt}. Given the input activations $X \in \mathbb{R}^{n \times d}$ for $n$ samples, we have:
\begin{equation}
H \approx \tfrac{1}{n}X^\top X + \lambda I, \qquad 
\end{equation}
where $\lambda > 0$ is a damping term used to improve the numerical condition and enhance the stability of the inversion process. Inspired by the second-order sensitivity idea in Optimal Brain Surgeon (OBS)~\cite{hassibi1993optimal}, the importance $S_{ij}$ of parameter $w_{ij}$ is defined as:
\begin{equation}
S_{ij} = \frac{w_{ij}^2}{[H^{-1}]_{jj}}.
\end{equation}

This measure indicates that parameter importance is related not only to the weight magnitude, but also to the local curvature on its corresponding input dimension. In practice, this process only requires collecting activation information through a single forward pass to complete the approximate estimations.

After obtaining the parameter importance estimates, we select the fraction $\eta$ of parameters with the lowest importance scores as the updatable subset, and during fine-tuning, only this parameter subset and the classification head are updated, while all other parameters remain frozen. Specifically, we introduce a binary mask matrix $M \in \{ 0, 1 \}$, where $M_{ij} = 1$ indicates that the corresponding parameter is allowed to be updated; otherwise, it remains frozen. The parameter update process can then be written as:
\begin{equation}
\theta \leftarrow \theta - \gamma \,(M \odot \nabla_{\theta}\mathcal{L}),
\end{equation}
where $\gamma$ is the learning rate, $\odot$ denotes element-wise multiplication, and $\mathcal{L}$ is the binary cross-entropy loss. In this way, CKI constrains the optimization within the subspace spanned by low-sensitivity parameters, allowing task knowledge to be injected into the model in a more controlled manner, thereby improving SID adaptation capability while minimizing damage to the pretrained representational structure as much as possible.

\begin{table*}[t]\small
\caption{Accuracy (ACC, \%) comparison of \ourmethod{} and other methods on the Setting 2. All methods are trained on ProGAN and evaluated on different test subsets. \textbf{Bold} indicates the best result, and \underline{underline} denotes the second-best.}
\centering
\scriptsize{
\resizebox{\linewidth}{!}{
\setlength\tabcolsep{2.2pt}
\renewcommand\arraystretch{1.15}
\begin{tabular}{l|ccccccccccccccccccc}
\hline\thickhline
\rowcolor{gray!20}
Method
& \rotatebox{55}{ProGAN}
& \rotatebox{55}{CycleGAN}
& \rotatebox{55}{BigGAN}
& \rotatebox{55}{StyleGAN}
& \rotatebox{55}{GauGAN}
& \rotatebox{55}{StarGAN}
& \rotatebox{55}{StyleGAN2}
& \rotatebox{55}{DDPM}
& \rotatebox{55}{PNDM}
& \rotatebox{55}{IDDPM}
& \rotatebox{55}{Guided}
& \rotatebox{55}{LDM\_200}
& \rotatebox{55}{LDM\_200\_cfg}
& \rotatebox{55}{LDM\_100}
& \rotatebox{55}{Glide\_100\_27}
& \rotatebox{55}{Glide\_50\_27}
& \rotatebox{55}{Glide\_100\_10}
& \rotatebox{55}{DALLE}
& \rotatebox{55}{Average}
\\
\hline

LGard
& 99.83 & 86.94 & 86.63 & 91.08 & 78.46 & 99.27 & 54.71 & 72.30 & 67.50 & 68.92 & 60.34 & 90.30 & 95.10 & 93.61 & 86.13 & 90.30 & 83.24 & 67.60 & 81.79
\\
\rowcolor{gray!10} DIRE
& 100.0 & 67.73 & 64.78 & 83.08 & 65.30 & 100.0 & 80.16 & 86.31 & 86.36 & 72.37 & 83.20 & 82.70 & 84.05 & 84.25 & 87.10 & 90.80 & 90.25 & 58.75 & 81.51
\\
NPR
& 99.75 & 71.90 & 90.10 & 94.30 & 77.70 & 100.0 & 87.80 & 88.00 & 82.30 & 83.80 & 68.35 & 89.80 & 90.10 & 89.05 & 76.55 & 80.15 & 79.50 & 85.40 & 85.25
\\
\rowcolor{gray!10} FreqNet
& 97.90 & 95.84 & 90.45 & 97.55 & 90.24 & 93.41 & 92.03 & 94.92 & 93.22 & 87.76 & 86.70 & 84.55 & 99.58 & 65.56 & 85.69 & 97.40 & 88.15 & 59.06 & 88.89
\\
FerretNet
& 99.90 & 98.80 & 92.60 & 98.0 & 91.40 & 99.10 & 98.50 & 88.82 & 94.90 & 86.00 & 86.40 & 98.8 & 98.5 & 98.8 & 97.3 & 97.2 & 97.9 & 91.4 & \underline{95.24}
\\
\rowcolor{gray!10} UniFD
& 99.90 & 98.40 & 94.65 & 84.95 & 99.40 & 96.75 & 83.35 & 71.50 & 86.15 & 73.35 & 69.65 & 94.40 & 74.00 & 95.00 & 78.50 & 79.05 & 77.90 & 87.30 & 85.79
\\
FatFormer
& 99.89 & 99.32 & 99.50 & 97.15 & 99.41 & 99.75 & 99.80 & 76.82 & 99.30 & 84.53 & 76.00 & 98.60 & 94.90 & 98.65 & 94.35 & 94.65 & 94.20 & 98.75 & 94.75
\\
\rowcolor{gray!10} RINE
& 100.0 & 99.30 & 99.60 & 88.90 & 99.80 & 99.50 & 97.82 & 78.90 & 96.08 & 89.20 & 76.10 & 98.30 & 88.20 & 98.60 & 88.90 & 92.60 & 90.70 & 85.00 & 92.64
\\
ForgeLens
& 99.95 & 99.24 & 97.67 & 96.64 & 98.84 & 95.24 & 96.20 & 81.23 & 98.44 & 89.42 & 73.34 & 98.72 & 96.98 & 98.86 & 96.07 & 96.17 & 95.43 & 98.29 & 94.82
\\
\rowcolor{gray!10} Effort
& 100.0 & 99.85 & 99.60 & 95.05 & 99.60 & 100.0 & 96.90 & 67.62 & 93.04 & 87.83 & 69.15 & 99.30 & 96.80 & 99.45 & 97.45 & 97.80 & 97.15 & 98.05 & 94.15
\\

\hline
\rowcolor[HTML]{D7F6FF}
\ourmethod{}
& 99.99 & 99.94 & 99.60 & 97.32 & 99.60 & 99.84 & 96.22 & 86.96 & 96.33 & 89.40 & 85.24 & 98.78 & 96.39 & 99.75 & 98.20 & 98.51 & 98.43 & 98.20 & \textbf{96.59}
\\
\hline\thickhline
\end{tabular}}}
\label{tab:table2}

\end{table*}

\section{Experiments}

\subsection{Experimental Setup}
\noindent\textbf{Datasets.}
To systematically evaluate our method, we follow widely adopted standard protocols in prior studies~\cite{chen2024single, tan2024rethinking} and conduct experiments under three settings. \textbf{Setting 1}: We train on the SDv1.4~\cite{rombach2022high} training set of GenImage~\cite{zhu2023genimage} and evaluate on the test sets corresponding to eight generators, covering mainstream diffusion models, GANs and large-scale online generators. \textbf{Setting 2}: We train on ProGAN~\cite{karras2017progressive} samples from ForenSynths~\cite{wang2020cnn} and evaluate on 18 generator-specific test sets collected from ForenSynths, DIRE~\cite{wang2023dire} and UniversalFake~\cite{ojha2023towards}, spanning GANs, diffusion models and diverse conditional generation configurations. \textbf{Setting 3}: We train separately on SDv1.4 or ProGAN, and test on two challenging benchmarks, Chameleon~\cite{yan2024sanity} and COSPY~\cite{cheng2025co}. 

\noindent\textbf{Compared baselines.}
We compare our method against 10 competitive baselines, including 5 specialized SID methods: LGard~\cite{tan2023learning}, DIRE~\cite{wang2023dire}, NPR~\cite{tan2024rethinking}, FreqNet~\cite{tan2024frequency} and FerretNet~\cite{liang2025ferretnet}, as well as 5 VFM-based detection methods: UniFD~\cite{ojha2023towards}, FatFormer~\cite{liu2024forgery}, RINE~\cite{koutlis2024leveraging}, ForgeLens~\cite{chen2025forgelens} and Effort~\cite{yan2024effort}. All baseline methods use either the optimal hyperparameter settings reported in the original papers or the officially released pretrained models.

\noindent\textbf{Evaluation metrics.}
We use classification accuracy (ACC) and average precision (AP) as the evaluation metrics. The model outputs are computed as scores with fake as the positive class and real as the negative, and the decision threshold for ACC is fixed at 0.5.

\noindent\textbf{Implementation details.}
In all experiments, we adjust images to a resolution of 224$\times$224 and apply only random horizontal flipping as data augmentation during training.
The original GenImage and ProGAN training sets contain approximately 324,000 and 144,000 images, respectively. During training, the compared methods follow their original training data scales, whereas our method is trained using only 1,600 sampled images from each set.
As the pretrained VFM, we adopt CLIP ViT-L/14~\cite{radford2021learning}, which is commonly used by prior baselines~\cite{yan2024effort, chen2025forgelens}, as the visual encoder to ensure comparability with existing VFM-based methods.
In CLI, only one epoch is used for critical layer identification.
In CKI, the damping term $\lambda$ for Hessian approximation is set to $1\mathrm{e}{-4}$ by default. For optimization, we use the Adam optimizer with $\beta_1=0.9$ and $\beta_2=0.999$, a batch size of 16, and an initial learning rate of $1\mathrm{e}{-4}$, which is decayed by a factor of 0.7 every 3 epochs. 

\begin{table}[t]
\centering
\caption{Accuracy (ACC, \%) comparison of \ourmethod{} and other methods on the Chameleon.}
\label{tab:table4}
\renewcommand{\arraystretch}{1.15}
\resizebox{\linewidth}{!}{%
\begin{tabular}{lccccc}
\hline\thickhline
\rowcolor{gray!15}
Training Dataset & NPR & FerretNet & ForgeLens & Effort & \ourmethod{} \\
\midrule
ProGAN          & 57.29 & 63.57 & \underline{87.64} & 84.13 & \textbf{88.70} \\
SDv1.4          & 58.13 & 67.92 & \underline{90.01} & 89.35 & \textbf{92.46} \\
ProGAN+SDv1.4   & 65.77 & 69.38 & \underline{90.32} & 89.98 & \textbf{93.07} \\
Average   & 60.40 & 66.96 & \underline{89.32} & 87.82 & \textbf{91.41} \\
\hline\thickhline
\end{tabular}%
}
\end{table}

\subsection{Comparison with Competing Methods}
\noindent\textbf{Setting 1.}
As shown in ~\cref{tab:table1}, in the cross-generator generalization evaluation on the GenImage benchmark, the test generators cover SD v1.4/v1.5~\cite{rombach2022high}, GLIDE~\cite{nichol2021glide}, VQDM~\cite{gu2022vector}, Wukong~\cite{wukong2022}, BigGAN~\cite{brock2018large}, ADM~\cite{dhariwal2021diffusion}, and Midjourney~\cite{midjourney2022}. Under this protocol, \ourmethod{} achieves an average ACC of 98.22\% and an average AP of 99.72\%, attaining the best overall performance, which demonstrates its stronger generalization capability in cross-distribution synthetic image detection. Meanwhile, most existing specialized detection methods exhibit more pronounced performance degradation on generators that differ substantially from the training distribution, particularly on test subsets such as BigGAN, VQDM, and ADM, reflecting their greater sensitivity to generator bias and distribution shifts. In contrast, \ourmethod{} shows more stable consistency across different test generators.

\begin{table}[t]
\centering
\caption{Accuracy (ACC, \%) comparison of \ourmethod{} and other methods on the COSPY. All methods are trained on SDv1.4.}
\label{tab:table5}
\renewcommand{\arraystretch}{1.15}

\resizebox{\linewidth}{!}{%
\begin{tabular}{lcccccc}
\hline\thickhline
\rowcolor{gray!15}
Methods &
\makecell{FLUX.1-\\dev} &
\makecell{PG-v2.5-\\1024} &
\makecell{SD-3-\\medium} &
\makecell{FLUX.1-\\sch} &
\makecell{SegMoE-\\SD} &
Average \\
\midrule
FerretNet & 94.80 & 94.13 & 79.85 & 94.40 & 96.87 & 92.01 \\
ForgeLens  & 95.93 & 96.21 & 85.86 & 86.00 & 100.0 & 92.80 \\
Effort    & 95.88 & 92.23 & 82.54 & 96.64 & 97.56 & \underline{92.97} \\
\midrule
\ourmethod{}       & 95.72 & 96.57 & 85.04 & 97.15 & 99.55 & \textbf{94.81} \\
\hline\thickhline
\end{tabular}%
}
\end{table}

\noindent\textbf{Setting 2.}
As shown in \cref{tab:table2}, this setting covers a variety of GANs and relatively early diffusion models.
The test generators are drawn from three benchmark suites: ProGAN, CycleGAN~\cite{zhu2017unpaired}, BigGAN, StyleGAN~\cite{karras2019style}, GauGAN~\cite{park2019semantic}, StarGAN~\cite{choi2018stargan} and StyleGAN2~\cite{karras2020analyzing} from ForenSynths. DDPM~\cite{ho2020denoising}, PNDM~\cite{liu2022pseudo}, and IDDPM~\cite{nichol2021improved} from DIRE. Guided Diffusion, DALLE~\cite{ramesh2021zero}, and LDM and GLIDE under different conditioning configurations from UniversalFake.
Under this setting, \ourmethod{} attains an average ACC of 96.59\%, improving by 1.35\% over the runner-up FerretNet (95.24\%).
Because this setting is trained on ProGAN, some methods tend to over-rely on GAN-specific artifact patterns and suffer pronounced distribution shift when transferred to the diffusion, leading to substantial drops on the DDPM and Guided test subsets and consequently lowering the overall average performance.
In contrast, \ourmethod{} maintains more consistent detection performance.

\noindent\textbf{Setting 3.}
As shown in \cref{tab:table4} and \cref{tab:table5}, we further evaluate more challenging cross-distribution benchmarks.
Chameleon better reflect real-world distributions, while COSPY evaluates models using a new generation of generators released after 2024, including SegMoE-SD~\cite{gupta2024progressive}, SD-3-medium~\cite{rombach2022high}, PG-v2.5-1024~\cite{li2024playground}, FLUX.1-sch~\cite{flux} and FLUX.1-dev.
In this setting, \ourmethod{} achieves the best performance on both Chameleon and COSPY, demonstrating stronger generalization to real distribution shifts and emerging generators.

\begin{table}[t]
\centering
\small
\setlength{\tabcolsep}{10pt}
\renewcommand{\arraystretch}{1.15}
\caption{Ablation study of CLI and CKI. We report average ACC performance (\%) under Setting~1 and Setting~2.}
\label{tab:ablation_cli_cki}
\resizebox{\linewidth}{!}{
\begin{tabular}{ccccc}
\hline\thickhline
\rowcolor{gray!15}
CLIP & w/ CLI & w/ CKI & Setting 1 & Setting 2 \\
\midrule
\ding{51} & \ding{55} & \ding{55} & 89.41 & 85.44 \\
\ding{51} & \ding{55} & \ding{51} & 95.12 & 94.61 \\
\ding{51} & \ding{51} & \ding{55} & 96.87 & 94.03 \\
\ding{51} & \ding{51} & \ding{51} & \textbf{98.22} & \textbf{96.59} \\
\hline\thickhline
\end{tabular}
}
\end{table}

\subsection{Ablation Studies}
\noindent\textbf{Effect of CLI and CKI module.}
To verify the contribution of each component in \ourmethod{}, we use the frozen CLIP visual encoder as the baseline and separately evaluate the variants with CKI only, CLI only, and the full model with both CLI and CKI. As shown in ~\cref{tab:ablation_cli_cki}, introducing either CLI or CKI alone consistently improves performance, indicating that the two components enhance the cross-distribution detection capability of SID from different perspectives. Specifically, CLI improves discriminative feature readout by identifying the critical representational layer that is more suitable for SID, whereas CKI enhances the stability of downstream adaptation by constraining the update range. When combined, the two achieve the best results, further demonstrating that CLI and CKI are complementary in \ourmethod{}.

\begin{figure}
    \centering
    \includegraphics[width=0.48\textwidth]{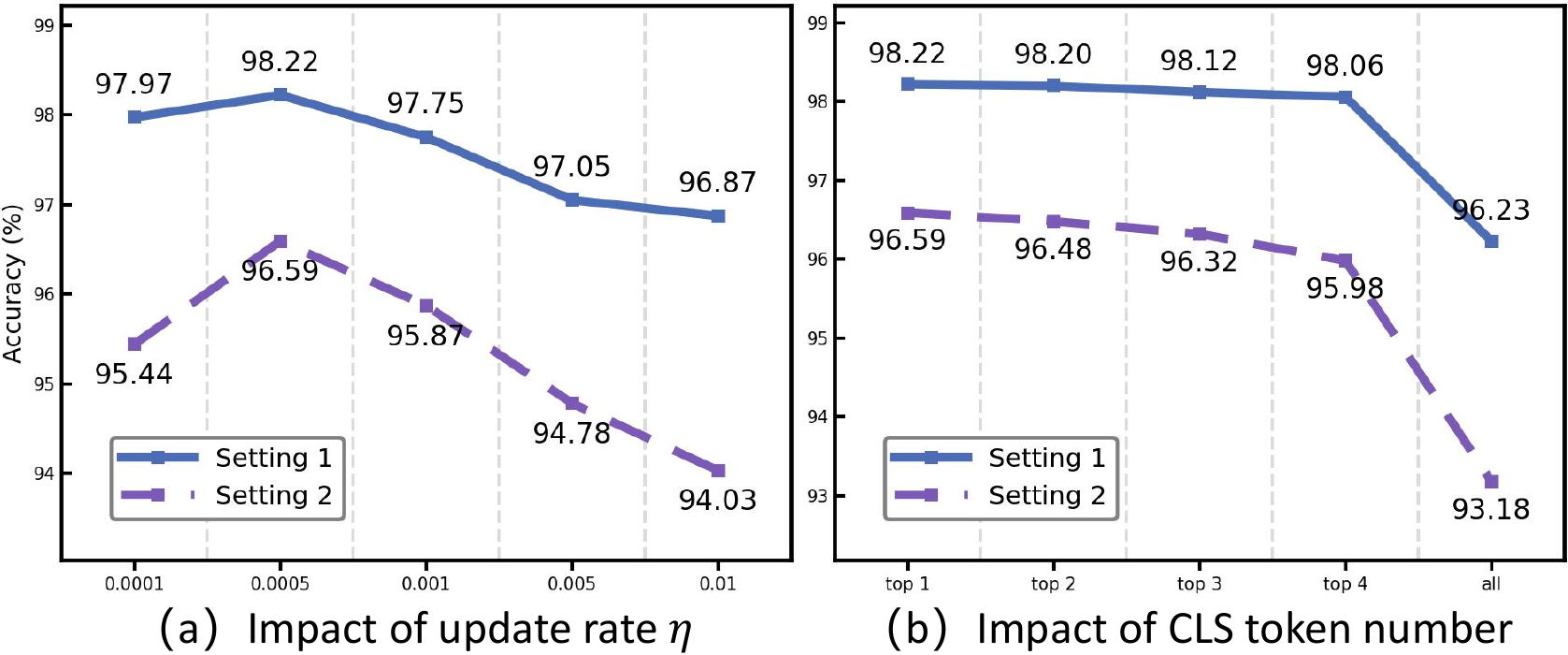}

    \caption{Impact of update rate $\eta$ and the number of CLS token used under Setting 1 and Setting 2.}
   
    \label{fig:ablation_hyper}
\end{figure}

\noindent\textbf{Effect of the update rate $\eta$.}
We gradually increase $\eta$ from a small ratio and evaluate the trend of average accuracy under Setting~1 and Setting~2. \cref{fig:ablation_hyper} (a) indicates that an overly small $\eta$ limits effective injection of forgery-related knowledge, whereas an overly large $\eta$ is more likely to induce representation drift and damage the pretrained structure, leading to degraded cross-distribution performance.
A stable optimum appears in the moderate range, with $\eta=0.0005$ performing best in both settings.

\noindent\textbf{Effect of the number of CLS tokens.}
We study the effect of using multiple CLS tokens for classification.
Specifically, we rank layers by the contributions learned in CLI and progressively construct a weighted feature set using $\{1,2,3,4,23\}$ CLS tokens.
The results are shown in \cref{fig:ablation_hyper} (b), as the number of included layers increases, the performance does not continue to improve. On the contrary, when the CLS tokens from all layers are used, the performance instead declines. This indicates that, for SID, simply fusing more hierarchical information does not lead to stronger generalization, but may instead dilute discriminative information or introduce irrelevant components that have undergone deep semantic compression.

\begin{figure}
    \centering
    \includegraphics[width=0.48\textwidth]{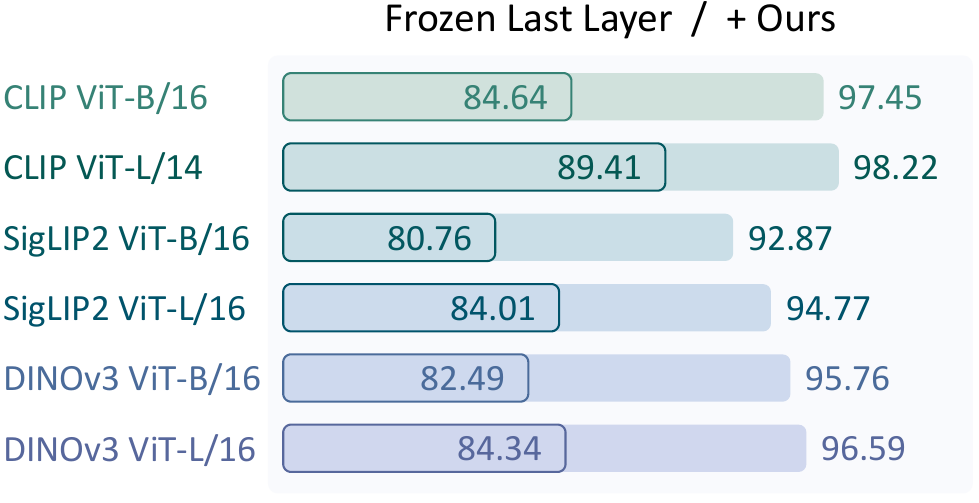}

    \caption{VFM backbones comparison. Performance of the frozen last-layer baseline and \ourmethod{} across different VFM.}
    \label{fig:VFM}
\end{figure}

\noindent\textbf{Generality across diverse VFM backbones.}
To validate the generality of \ourmethod{} across VFM architectures, we reproduce experiments on a range of mainstream vision foundation models, including CLIP ViT-L/14, CLIP ViT-B/16, SigLIP2 ViT-B/16~\cite{tschannen2025siglip2multilingualvisionlanguage}, SigLIP2 ViT-L/16, DINOv3 ViT-B/16~\cite{simeoni2025dinov3}, and DINOv3 ViT-L/16.
When switching backbones, all other training settings are kept unchanged.
The results in \cref{fig:VFM} show that \ourmethod{} consistently yields performance gains across architectures and model scales, suggesting that critical layer identification and controlled knowledge injection do not rely on a specific pretrained model or a fixed layer hierarchy, but instead constitute a transferable SID adaptation strategy.

\section{Conclusion}

To address insufficient cross-distribution generalization in synthetic image detection, this paper systematically analyzes the hierarchical representational properties of pretrained vision foundation models in the SID, as well as the pretrained structures underlying their generalization capability. On this basis, we propose the \ourmethod{} framework. On the one hand, it selects intermediate-layer representations more effective for forgery discrimination through adaptive critical layer identification. On the other hand, it restricts downstream updates to a low-sensitivity parameter subspace through controlled knowledge injection, thereby improving SID discrimination while minimizing disturbance to pretrained representations and stable inductive biases. Extensive cross-generator experiments show that \ourmethod{} achieves consistent performance improvements under few-shot training conditions while maintaining good transferability across different VFM architectures and model scales, validating its generalization capability and versatility in cross-domain SID tasks.

\textit{Limitation.} Although \ourmethod{} shows strong generalization in cross-distribution SID, it has several limitations. First, the design of \ourmethod{} is grounded in empirical observations and currently lacks a rigorous theoretical justification. Second, our experiments are primarily focused on image-level detection, and its applicability to finer-grained forensic scenarios has yet to be thoroughly investigated.

\section{Acknowledgments}
This work is supported by Taobao \& Tmall Group of Alibaba through Alibaba Research Intern Program and Alibaba Group Innovative Research Program, CCF-Taobao \& Tmall Group TECH Kangaroo Fund (No. CCF-Taobao \& Tmall Group 2026002), Zhejiang Provincial Natural Science Foundation of China under Grant No. LMS26F020019 and Information Technology Center and State Key Lab of CAD\&CG, Zhejiang University.

\bibliographystyle{ACM-Reference-Format}

\end{document}